\documentclass[11pt]{article}
\usepackage[table]{xcolor}

\usepackage[preprint]{acl}

\usepackage{times}
\usepackage{latexsym}
\usepackage{multirow}
\usepackage{amssymb}
\usepackage{booktabs}
\usepackage{amsmath}
\usepackage{enumitem}
\usepackage{graphicx}
\usepackage{float}

\usepackage{tcolorbox}
\tcbuselibrary{skins, breakable}

\newtcolorbox{promptbox}[2][]{
  enhanced,
  breakable,
  colback=white,
  colframe=black!70,
  colbacktitle=black!70,
  coltitle=white,
  title={#2},
  fonttitle=\bfseries,
  left=6pt, right=6pt, top=6pt, bottom=6pt,
  boxrule=0.6pt,
  arc=2mm,
  #1
}

\usepackage[T1]{fontenc}

\usepackage[utf8]{inputenc}

\usepackage{microtype}

\usepackage{inconsolata}

\usepackage{graphicx}
\usepackage{pifont}
\usepackage{tikz,xcolor}

\definecolor{SectionGray}{HTML}{ECEFF1}
\definecolor{LightBlue}{HTML}{F0F8FF}
\definecolor{lightyellow}{HTML}{FFF8BA}
\title{OmniMem: Perturbation-aware Memory Compression for Streaming Audio-Visual LLMs}

\author{
\textbf{Guangzhi Sun\textsuperscript{2,3}},
 \textbf{Yixuan Li\textsuperscript{1,2}},
 \textbf{Yudong Yang\textsuperscript{1,2}}
  \textbf{Chao Zhang\textsuperscript{1}}
\\
\\
 \textsuperscript{1}Tsinghua University \\
 \textsuperscript{2} ByteDance \\
 \textsuperscript{3} Department of Engineering, University of Cambridge
\\
 \texttt{gs534@cam.ac.uk}
}

\begin{document}
\maketitle
\begin{abstract}
Audio-visual large language models (LLMs) hold strong promise for long-form video understanding, yet their long-video inference is fundamentally limited by the linear growth of video tokens and key-value (KV) caches. We present OmniMem, a memory-efficient streaming framework designed specifically for audio-visual LLMs.
Unlike existing compression methods that treat all tokens uniformly, OmniMem introduces a modality-aware memory allocation strategy that separately manages visual and audio contexts, addressing the severe token imbalance between the two modalities. OmniMem further preserves informative and non-redundant KV states through perturbation-aware memory selection, enabling compact memory without sacrificing long-range understanding. To strengthen compression under realistic deployment constraints, we also explore budget-aware fine-tuning, which encourages the model to consolidate useful information into retained memory. Experiments on VideoMME Long, LVBench, and LVOmniBench with video-SALMONN 2+ and Qwen-2.5-Omni show that OmniMem consistently improves over strong training-free compression baselines by 2–4\% absolute accuracy under the same memory budgets, with an additional 1–2\% gain after fine-tuning.\footnote{\url{https://github.com/bytedance/SALMONN/tree/omni_mem}}

\end{abstract}

\section{Introduction}

Audio-visual large language models (av-LLMs) are increasingly used for long video understanding. However, the number of video tokens or key-value (KV) cache grows linearly with video length, quickly reaching the GPU memory limit in streaming inference for hour-long videos. Chunked streaming processing, which processes video token sequence chunk-by-chunk and compresses either the video tokens or the KV cache after each chunk, has emerged as the practical solution. On one hand, compression methods have been applied to the video input tokens, such as PEMF \cite{pemf} and video-SALMONN S \cite{vss}, where complex memory structures were adopted to maintain a fixed memory budget. On the other hand, recent work including StreamMem~\cite{streammem}, StreamingKV~\cite{streamkv}
InfiniPot-V~\cite{infinipotv}, and HERMES~\cite{hermes} maintain a fixed budget for KV cache. Both research directions demonstrate that significant compression is achievable without catastrophic quality loss, establishing chunked streaming as the \textit{de facto} paradigm for memory-constrained long video inference.

Within the KV cache compression landscape, two design dimensions have received increasing attention. The first is per-layer budget allocation. A substantial body of work, including PyramidKV \cite{pyramid}, SqueezeAttention \cite{squeezed}, LAVa \cite{lava}, and EvolKV \cite{evolkv}, has established that non-uniform layer budgets consistently outperform uniform allocation. However, these methods operate offline on the full sequence and do not transfer directly to the streaming setting, where only chunk-local signals are available. The second is token selection within a given budget. Existing streaming video methods measure token redundancy by cosine similarity and score token importance by attending over the KV cache with a fixed generic prompt. 
Existing methods typically estimate token importance using a generic proxy query. However, because the proxy query is external to the model’s computation graph, its attention scores may not accurately reflect the perturbation caused by token eviction. Therefore, low proxy attention does not guarantee safe eviction.
Furthermore, with both audio and visual inputs, a unified token selection metric with the generic proxy query (often derived from visual language templates) may systematically under-weigh audio tokens regardless of their contribution to the model's output.

In this paper, we address the limitations in both dimensions with OmniMem, a perturbation-aware, audio-visual memory compression framework for streaming long video understanding, particularly focusing on videos spanning multiple hours. Specifically, OmniMem combines the attention-based importance score with the cosine similarity-based redundancy metric to select KV pairs to retain. That is, the selected tokens are to minimize the distortion to the attention output after eviction. For budget allocation, we show that layer compressibility in av-LLMs is governed jointly by normalized attention entropy and intra-layer value cosine similarity,
and adopt a per-layer per-modality KV cache budget scheme in OmniMem. Moreover, OmniMem uses an Audio-Visual Budget Allocation (AVBA) to ensure that compression signals are comparable across the large visual-to-audio token imbalance, preventing the audio modality from being treated as simply a small group of visual tokens. Furthermore, we show that fine-tuning of the LLM backbone with the layerwise budget and suitable gradient back-propagation design helps the model to consolidate information into retained tokens, providing a further gain beyond the training-free configuration.

Experiments on Video-SALMONN~2+~\cite{videosalmonn2} (4B and 8B) and
Qwen2.5-Omni~\cite{qwen25omni} across three long video understanding benchmarks involving both audio and visual inputs, including VideoMME Long~\cite{videomme},
LVBench~\cite{lvbench}, and LVOmniBench~\cite{lvomnibench}. Consistent 2--3\% absolute accuracy improvements were obtained over generic proxy query-based methods with the same memory budgets without training. A further consistent 1--2\% absolute accuracy improvement was obtained after fine-tuning. The main contributions of this paper are summarised as follows.

\begin{itemize}[leftmargin=*]
    \item We propose OmniMem, a perturbation-aware modality-aware KV Cache selection mechanism for streaming long video understanding. OmniMem combines attention-based importance with cosine-similarity-based redundancy, minimizing the perturbation to the output after eviction.
    \item In OmniMem, we propose the first audio-visual budget allocation mechanism, AVBA, for KV cache memory for av-LLMs. AVBA addresses the modality representation discrepancies that are not reflected in any of the metrics used for KV cache compression before.
    \item OmniMem demonstrated consistent 2-3\% absolute accuracy improvements compared to cosine-similarity and generic prompt baselines, and can be further improved by 1-2\% when finetuned with allocated audio and visual budgets. 
\end{itemize}

\section{Related Work}

\subsection{Streaming long video understanding}
Recent progress in long video understanding has shifted toward streaming inference, where inputs are processed chunk-by-chunk under strict memory constraints. Early approaches rely on frame sampling or token pooling, which are insufficient for long-duration videos. More recent systems adopt bounded-memory streaming. One mainstream of research investigates training-free KV cache compression \cite{rekv,streammem,infinipotv,streamingtom,streamkv,hermes}. These methods usually adopt cosine-similarity-based metrics to reduce redundancy at the input, and then use a generic proxy query, often a generation template, to compute attention scores, which are then used to represent the importance of each KV pair. Instead of KV cache compression, a series of papers explores the Transformer-XL structure for long video understanding \cite{shu2024video,liu2025video}. Another stream of work builds memory modules at the input to the backbone LLM \cite{pemf,longvila,moviechat,longva,longvu}, or leveraging a combination of input memory structure and KV cache selection \cite{vss}. Despite their effectiveness, most methods rely on heuristic token scoring and do not explicitly optimize the impact of eviction on model outputs.

\subsection{KV cache budget in long-context LLMs}
Beyond token selection, recent works highlight the importance of non-uniform layer-wise allocation. PyramidKV~\cite{pyramid}, SqueezeAttention~\cite{squeezed}, LAVa~\cite{lava}, and EvolKV~\cite{evolkv} show that allocating different budgets to different layers considerably improves performance. These methods are primarily developed for offline settings based on global sequence statistics, rendering them less effective for streaming scenarios constrained by local context.


\subsection{Audio-visual LLMs}
For audio-visual LLMs, compression is further complicated by modality imbalance and heterogeneous token characteristics. Existing methods generally apply unified token scoring mechanisms, often based on cosine similarity or attention from generic proxy queries, which can bias selection toward visually dominant tokens. Models such as Video-SALMONN~2+~\cite{videosalmonn2} and Qwen2.5-Omni~\cite{qwen25omni} highlight the importance of joint audio-visual reasoning, but modality-aware compression remains underexplored. Some recent works investigate cross-modal alignment and adaptive token pruning in multimodal transformers, yet they do not address KV cache compression explicitly. In contrast, our approach introduces perturbation-aware token selection and modality-specific budget normalization, enabling more balanced and principled compression in streaming av-LLMs.

\begin{figure*}[t]
    \centering
    \includegraphics[width=0.95\linewidth]{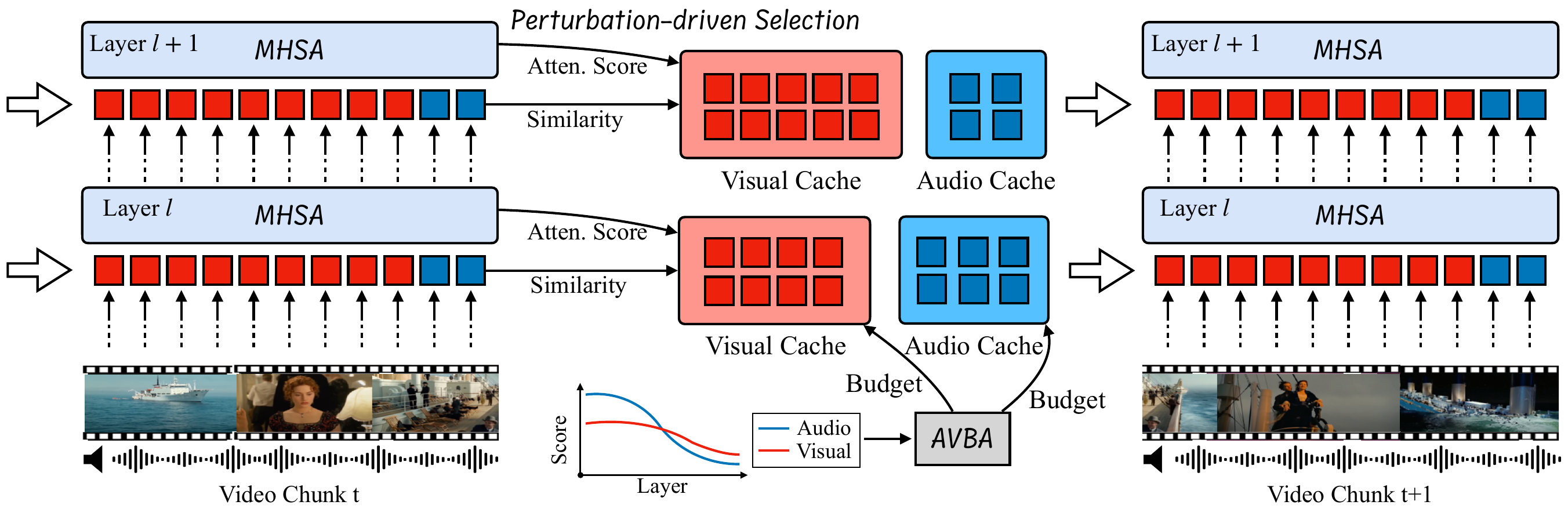}
    \vspace{-0.3cm}
    \caption{The illustration of the OmniMem mechanism. The video is processed chunk by chunk. For each input chunk, multi-head self-attention (MHSA) uses the past KV cache and the input of the current chunk. The KV cache is updated with new pairs from the current chunk, and then pruned to a fixed size based on attention score and similarity (Section \ref{sec:selection}). KV pairs corresponding to audio and visual positions are separately stored in two caches, with each budget determined by AVBA (Section \ref{sec:budget}).}
    \label{fig:pmamem}
\end{figure*}

\begin{figure*}[h]
    \centering
    \includegraphics[width=\linewidth]{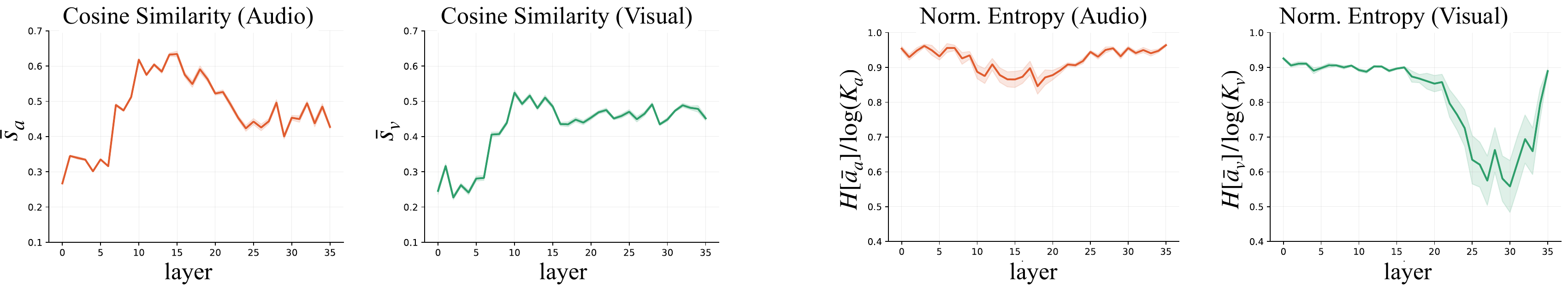}
    \vspace{-0.5cm}
    \caption{Plot of cosine similarity and normalized entropy against layers for audio and visual positions on a small calibration set. Shaded areas represent one standard deviation measured across all chunks in all videos.}
    \label{fig:avbascore}
\end{figure*}

\section{Methodology}

The workflow for the OmniMem mechanism is illustrated in Fig. \ref{fig:pmamem}. During the prefill stage, the video is processed chunk by chunk, with each chunk containing a specific number of audio and visual tokens. For layer $l$ at chunk $c$, denoting the input hidden states as ${H}_c^{(l)}$, the multi-head self-attention (MHSA) is computed as
\begin{equation}
    Y_c^{(l)} = \text{MHSA}(Q_c^{(l)},K_c^{(l)},V_c^{(l)}, \mathcal{O}_{c}^{(l)}, \mathcal{P}_{c}^{(l)}),
    \label{eq:mhsa}
\end{equation}
where $Q_c,K_c^{(l)},V_c^{(l)}$ are computed by linear projections from $H_c^{(l)}$, and $\mathcal{O}_{c}^{(l)}, \mathcal{P}_{c}^{(l)}$ denote visual and audio KV cache respectively, which are two separate KV caches to store KV pairs corresponding to audio and visual token positions as shown below:
\begin{equation}
    \mathcal{O}_{c}^{(l)} = (\bar{K}_{c,v}^{(l)}, \bar{V}_{c,v}^{(l)}), \quad \mathcal{P}_{c}^{(l)} = (\bar{K}_{c,a}^{(l)}, \bar{V}_{c,a}^{(l)}),
\end{equation}
where subscript $a$ and $v$ denotes KV pairs corresponding to audio and visual token positions respectively. After computing MHSA, the KV cache is updated with the pairs of the current chunk by
\begin{align}
    \tilde{K}_{c+1,v}^{(l)} &= [\bar{K}_{c,v}^{(l)}; {K}_{c,v}^{(l)}],~ \tilde{V}_{c+1,v}^{(l)} = [\bar{V}_{c,v}^{(l)}; {V}_{c,v}^{(l)}] \\
    \tilde{K}_{c+1,a}^{(l)} &= [\bar{K}_{c,a}^{(l)}; {K}_{c,a}^{(l)}],~ \nonumber\tilde{V}_{c+1,a}^{(l)} = [\bar{V}_{c,a}^{(l)}; {V}_{c,a}^{(l)}].
\end{align}
KV cache is then pruned to a fixed size by selecting KV pairs to retain, as shown below:
\begin{align}
    \mathcal{I}_{c,v}^{(l)} &= \text{Select}(\tilde{K}_{c+1,v}^{(l)}, \tilde{V}_{c+1,v}^{(l)}, M_v) \\
    \mathcal{I}_{c,a}^{(l)} &= \text{Select}(\tilde{K}_{c+1,a}^{(l)}, \tilde{V}_{c+1,a}^{(l)}, M_a).
    \label{eq:select}
\end{align}
where $\mathcal{I}_{c,v}^{(l)}, \mathcal{I}_{c,a}^{(l)}$ are indices to keep for visual and audio positions, and $M_v, M_a$ are cache budgets for visual and audio KV cache. The selection is based on both cosine similarity (redundancy) and attention scores (importance) as described in Section \ref{sec:selection}. The budget for KV cache given to audio and visual token positions is different for different layers in general, which is determined by the AVBA module described in Section \ref{sec:budget}. 

\subsection{Perturbation-aware KV Cache Selection}
\label{sec:selection}
Existing KV cache selection methods mostly uses attention scores from a guidance prompt \cite{streammem,hermes}. Instead of relying on a hand-crafted guidance prompt, we adopt a perturbation-aware selection criterion, which evicts KV cache entries that have minimal influence to the attention output of the current chunk.

Specifically, when a pair $k$ is dropped from the KV cache, the attention output changes for two independent reasons. First, the attention mass that was directed to pair $k$ is redistributed over the retained pairs. If pair $k$ received a large attention weight, this redistribution is large. Second, even after redistribution, the output changes because the retained pairs may not carry the same information as the dropped pair. If pair $k$'s value vector pointed in a direction not represented by the retained pair, the redistributed attention cannot recover that information. A KV pair is therefore expensive to drop only when both conditions hold simultaneously. 

We compute the first condition using the accumulated attention score on each KV pair, representing the importance of each pair is:
\begin{equation}
    a_k = \sum_{q\in|Q|} A_{q,k},
    \label{eq:accum}
\end{equation}
where $A_{q,k}$ is the attention score from query index $q$ to key index $k$. We omit the layer superscript and chunk subscript here for simplicity\footnote{In practice, we can compute the attention score by splitting the query into 256-vector groups to keep the same level of memory consumption as the guidance prompt.}. The second condition can be represented by the cosine similarity of the KV pair $k$ with their adjacent positions, computed using hidden states by
\begin{equation}
    s_k = [\cos (H_{k}, H_{k+1}) + \cos (H_{k-1}, H_{k})]/2.
    \label{eq:cossim}
\end{equation}
Note that this computation requires retaining hidden states instead of the KV cache, which causes slight overhead to memory and inference time. To maintain the same KV cache setting as other KV cache methods, similarity can be computed from value vectors. The selection criterion is the combination of both conditions, as shown below:
\begin{equation}
    \psi_k = a_k^{\lambda}{(1 - s_k)},
    \label{eq:psi}
\end{equation}
where the hyper-parameter $\lambda$ is a normalization factor since the range of $a_k$ and $s_k$ can be very different. The select function thus selects the top $K$ pairs that have the highest score $\psi$ to retain. The attention weight $a_k$ determines the importance, where a higher value means a vector is more frequently used by the subsequent layer. The cosine similarity $s_k$ reflects the redundancy. The multiplicative combination means that we keep non-redundant and mostly-used tokens in memory.

\subsection{Audio-Visual Budget Allocation}
\label{sec:budget}


In av-LLMs, visual tokens are often 30-40 times more than audio tokens. When a uniform budget is applied, only a small portion of audio tokens can be retained, potentially discarding important tokens corresponding to human speech, which is less compressible than visual tokens. Therefore, we propose AVBA to allocate separate budgets for the KV cache for audio and visual token positions.


To determine KV cache budgets for audio and visual tokens, we first define a fixed prior visual to audio token ratio $r$ which is a hyper-parameter for a model depending on the nature of the audio and visual encoders, and we fix this to be 5 throughout this paper. Then, we allow fluctuations of budgets around this ratio by measuring the compressibility using normalized entropy and cosine similarity by
\begin{align}
    \mathcal{C}_v &= H[\bar{a}_v](1 - \bar{s}_v) / \log(K_v), \\
    \mathcal{C}_a &= H[\bar{a}_a](1 - \bar{s}_a) / \log(K_a),
\end{align}
where $\mathcal{C}$ are compressibility scores for audio and visual KV cache, $\bar{a}_v, \bar{a}_a$ denote the attention probabilities from Eqn. \eqref{eq:accum} normalized on visual and audio tokens, and $\bar{s}_v, \bar{s}_a$ denote average cosine similarity on visual and audio tokens. We omit the layer superscript and the chunk subscript for simplicity. An example plot for both measures are shown in Fig. \ref{fig:avbascore}, where audio and visual positions show different patterns. $K_v$ and $K_a$ are the number of visual and audio tokens before eviction to normalize the entropy. The audio and visual proportions are given below:
\begin{align}
    w_v &= (\mathcal{C}_v * r) / (\mathcal{C}_v * r + \mathcal{C}_a),\\
    w_a &= \mathcal{C}_a / (\mathcal{C}_v * r + \mathcal{C}_a).
    \label{eq:av_alloc}
\end{align}
The budget is then split according to $w_v$ and $w_a$. We adopt a non-uniform budget based on overall compressibility across layers. For layer $l$, there is
\begin{equation}
    \mathcal{C}^{(l)} = H[\bar{a}^{(l)}](1 - \bar{s}^{(l)}).
\end{equation}
The total budget $\mathcal{B}^{(l)}$ allocated to layer $l$ becomes 
\begin{equation}
    \mathcal{B}^{(l)} = \text{Softmax}(\mathcal{C}^{(l')}/T)[l],
    \label{eq:alloc}
\end{equation}
where the temperature hyper-parameter $T$ controls the degree of budget variation across layers. In practice, the budget per layer is constrained within the range of $[\mathcal{B}_\text{min}, \mathcal{B}_\text{max}]$. The budget is determined offline for each model based on a small calibration set of videos, and there is no need for any user prompts to determine the budget.

\subsection{Fine-tuning with Allocated Budgets}

\begin{figure}[t]
    \centering
    \includegraphics[width=0.95\linewidth]{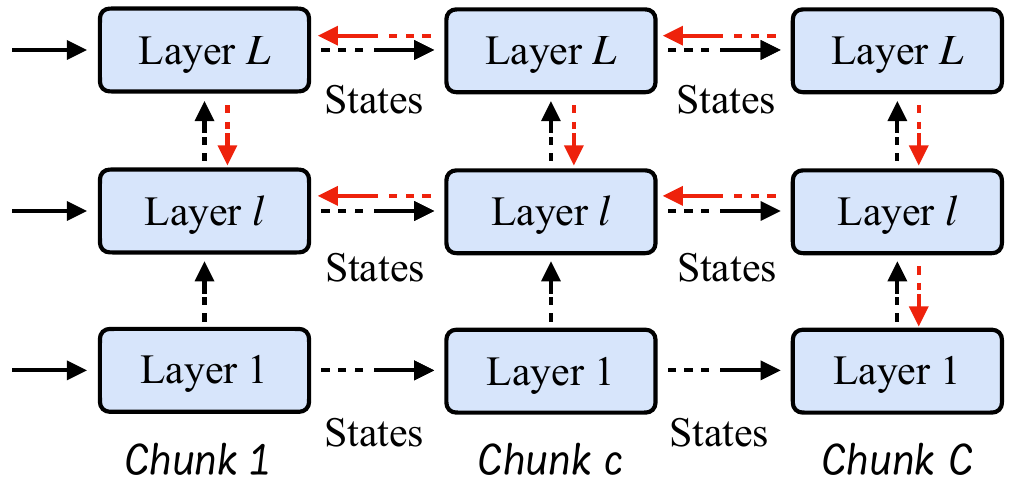}
    \caption{Fine-tuning process with allocated budgets. Red arrows are the gradient paths. States are compressed, hidden states carried over from the last chunk.}
    \label{fig:finetune}
\end{figure}

While OmniMem is primarily proposed as a training-free method, we investigate whether supervised fine-tuning with the allocated budget can further improve the performance. Specifically, the fine-tuning process is shown in Fig. \ref{fig:finetune}, where the same chunked video processing is applied. Since the aim is to fit the budget allocation, we carry hidden states instead of KV cache with just a cosine-similarity based selection criterion for low computational cost and easier implementation. Furthermore, when the sequence is too long to afford a full-sequence back-propagation, we truncate the back-propagation to the last $L-l$ layers while ensuring the gradient flows through all the layers at the last chunk, as illustrated in Fig. \ref{fig:finetune}.

\begin{table*}[t]
    \centering
    \caption{Accuracies (\%) using 3 different models on long video understanding benchmarks. All systems operate at 1FPS and retain 8K KV cache entries. Uniform denotes a uniform index selection for the KV cache. For video-SALMONN 2+, we also include results by fine-tuning with an allocated memory budget (denoted as +SFT).}
    \resizebox{\linewidth}{!}{
    \begin{tabular}{llccc}
    \toprule
    \textbf{Models} & \textbf{Methods}     & \textbf{Video-MME long} & \textbf{LVBench} & \textbf{LV-Omni-Bench} \\
    \midrule
    \multirow{5}{*}{video-SALMONN 2+ (8B)} & Uniform & 64.6 & 47.4 & 38.8 \\
     & InfiniPot-V    & 64.5 & 47.7 & 39.0 \\
     & StreamMem    & 65.1 & 50.3 & 40.5 \\
    & HERMES    & 65.0 & 49.8 & 40.2 \\
    & OmniMem (ours)    & \textbf{69.6} & \textbf{53.3} & \textbf{42.5} \\
    \rowcolor{LightBlue} & OmniMem + SFT (ours) & \textbf{70.2} & \textbf{55.7} & \textbf{43.1} \\
    \midrule
    \multirow{5}{*}{video-SALMONN 2+ (4B)}   & Uniform  & 59.9 & 46.1 & 36.6 \\
    & InfiniPot-V  & 60.6 & 46.8 & 37.5 \\
    & StreamMem    & 61.9 & 47.1 & 37.5 \\
    & HERMES    & 61.6 & 47.3 & 37.2 \\
    & OmniMem (ours)    & \textbf{64.4} & \textbf{50.3} & \textbf{39.8} \\
    \rowcolor{LightBlue} & OmniMem + SFT (ours)    & \textbf{65.2} & \textbf{54.4} & \textbf{40.7} \\
    \midrule
    \multirow{5}{*}{Qwen-2.5-Omni (7B)} & Uniform & 50.1 & 37.1 & 32.6 \\
    & InfiniPot-V    & 50.3 & 37.3 & 32.9 \\
    & StreamMem    & 49.8 & 37.6 & 32.8 \\
    & HERMES    & 50.5 & 37.5 & 32.8  \\
    & OmniMem (ours)    & \textbf{51.9} & \textbf{38.8} & \textbf{34.3} \\
    \bottomrule
    \end{tabular}}
    \label{tab:main}
\end{table*}

\section{Experimental Setup}

\subsection{Data}

We focus our experiments on three general long video understanding benchmarks and one streaming benchmark that require both audio and visual as inputs. For general long video understanding, we evaluate on VideoMME long partition \cite{videomme}, LVBench \cite{lvbench} and LV-Omni-Bench \cite{lvomnibench}, where LVBench and LV-Omni-Bench contain videos over 2 hours. For streaming baselines, we evaluate on StreamingBench \cite{streamingbench} which contains both a real-time perception partition and an omni-source partition, testing both real-time perception ability and the comprehension combining both audio and visual modalities.
Moreover, fine-tuning with allocated budget uses the official supervised fine-tuning (SFT) data in video-SALMONN 2+ \cite{videosalmonn2}.

\subsection{Models and Baseline Methods}
We focus our investigation on dense models. Specifically, we closely follow video-SALMONN 2+ \cite{videosalmonn2} training pipeline and fine-tune with Qwen3-VL-4B and -8B backbones. We also use Qwen2.5-Omni \cite{qwen25omni} 7B model as another audio-visual mode with a different structure to show generalizability.

For baseline methods, we primarily compare with training-free KV cache compression, including InfiniPot-V \cite{infinipotv}, StreamMem \cite{streammem} and HERMES \cite{hermes}. We use the best hyper-parameter setting reported in HERMES paper directly.

\subsection{Inference Specifications}

Inference is performed at 1FPS with a fixed 360p resolution. If not specified, the default memory size is 8K KV pairs per layer, or equivalent when a variable layerwise budget is used. For practicality, we do not set any maximum number of frames since no one would know when the video would stop. This means in the extreme cases, the total number of input tokens could approach 1M. 

Regarding hyper-parameter settings, without specification, we default $\lambda$ in Eqn. \eqref{eq:psi} to be 0.02, and $T$ in Eqn. \eqref{eq:alloc} to be 0.2. The budget distribution under this temperature is shown in Appendix \ref{sec:distribution}. For fine-tuning with the allocated budget, we set gradient back-propagation to be cut off at the middle layer, which is 18 for video-SALMONN 2+, and fine-tune for 2 epochs. We only fine-tune video-SALMONN 2+ because it provides the complete fine-tuning recipe. All experiments are conducted on H800 GPU machines. For the video-SALMONN 2+ (8B) model, inference on a single H800 GPU takes 9 hours for VideoMME-long, 32 hours for LVBench and 12 hours for LV-Omni-Bench. Fine-tuning with the allocated budget is performed on 32$\times$H800 for 36 hours. Complete implementation code as well as fine-tuned model checkpoints will be released.

\section{Results}

\subsection{Main Results}

\begin{table}[t]
    \centering
    \caption{Accuracies on StreamingBench using video-SALMONN 2+ (8B) and (4B), and Qwen-2.5-Omni (7B). StreamingBench results are given as Realtime, Omni-source and Contextual partitions (R/O/C).}
    \resizebox{\linewidth}{!}{
    \begin{tabular}{lccc}
    \toprule
    Methods     &  Realtime & Omni-source & Contextual \\
    \midrule
    \rowcolor{SectionGray} \multicolumn{4}{l}{video-SALMONN 2+ (8B)} \\
    StreamMem   & {77.6} & 56.5  & 39.7 \\
    HERMES & {77.9} & 57.8 & 40.1  \\ 
    OmniMem (ours) & \textbf{78.5} & \textbf{60.9} & \textbf{40.7} \\
    \rowcolor{SectionGray} \multicolumn{4}{l}{video-SALMONN 2+ (4B)} \\
    StreamMem   & 77.1 & 51.6 & 35.7 \\
    HERMES & 77.3 & 51.7 & 36.6 \\ 
    OmniMem (ours) & \textbf{77.5} & \textbf{58.9} & \textbf{37.7} \\
    \rowcolor{SectionGray} \multicolumn{4}{l}{Qwen-2.5-Omni (7B)} \\
    StreamMem   & 68.1 & 35.0 & 33.9 \\
    HERMES & 68.2 & 34.5 & \textbf{34.5} \\ 
    OmniMem (ours) & \textbf{68.4} & \textbf{38.4} & {34.2} \\
    \bottomrule
    \end{tabular}}
    \label{tab:stream}
\end{table}

Results on the 3 long video understanding benchmarks are shown in Table \ref{tab:main}, and on streaming benchmarks are shown in Table \ref{tab:stream}. On long video understanding, OmniMem achieved substantially better performance than the baseline KV cache compression methods. With the video-SALMONN 2+ (8B) model, compared to the best performing KV cache compression such as HERMES and StreamMem, OmniMem achieved 4.5\%, 3.0\% and 2.2\% absolute accuracy improvements on video-MME long partition, LVBench and LV-Omni-Bench, respectively. Similar levels of improvements were found on video-SALMONN 2+ (4B) and Qwen-2.5-Omni (7B), with the overall improvements being slightly more modest on Qwen-2.5-Omni due to limited audio modeling ability.

Performance improvements were also observed on streaming understanding tasks, especially on audio-visual understanding. On StreamingBench, OmniMem achieved slightly better performance on the real-time partition since it is mainly optimizing memory rather than real-time perception. More importantly, notably larger improvements were observed on the omni-source understanding partition, where a better audio-visual balance is necessary to achieve better performance. As a result, 3.1\% and 7.1\% absolute accuracy improvements were obtained on the Omni-source partition using video-SALMONN 2+ (8B) and (4B), respectively. Furthermore, we provide ablation studies on each important components in Appendix \ref{sec:ablation}.

\textbf{Fine-tuning with allocated budget further improved} both 8B and 4B video-SALMONN 2+ models. The largest improvement happened on LVBench, with 2.4\% and 4.1\% for 8B and 4B models respectively. During training, LLM learns to extract and leverage the information stored in the fixed memory budget more effectively and efficiently. LVBench has longer videos than the other two datasets, which requires more memory capacity. This trend is also reflected in Fig.~\ref{fig:memvar}, where performance on LVBench continues to benefit from increased memory budgets, while the other two datasets become saturated earlier. As a result, improvements in memory utilization and information consolidation provide particularly strong benefits on LVBench.





\subsection{Influence of Audio-visual Token Ratio}

\begin{table}[t]
    \centering
        \caption{Ablation studies on the ratio between visual and audio memory, $r$ in Eqn. \eqref{eq:av_alloc} of OmniMem using video-SALMONN 2+ (8B) model. No split denotes using only a single cache rather than separate audio and visual ones, which is equivalent to $r$ around 40-50.}
    \resizebox{\linewidth}{!}{
    \begin{tabular}{lcccc}
    \toprule
    Methods     & VMME long & LVBench & LV-Omni-Bench  \\
    \midrule
    No Split & 66.8 & 52.2 & 40.7 \\
    $r=20$     & 66.7 & 52.7 & 41.4 \\
    $r=10$ & 68.6 & 53.4 & 41.6 \\
    $r=5$ & 69.6 & 53.3 & 42.5  \\
    $r=2$ & 68.8 & 51.5 & 43.2 \\
    \bottomrule
    \end{tabular}}
    \label{tab:ratio}
\end{table}

We conduct ablation studies on the ratio between visual and audio memory, i.e. $r$ in Eqn. \eqref{eq:av_alloc} of OmniMem using video-SALMONN 2+ (8B) model as shown in Table \ref{tab:ratio}. At a high ratio, e.g. $r=20$ which is close to the original model allocation without separate memories, the performance is also close, with performance on VideoMME long partition and LV-Omni-Bench dropping significantly down by around 2-3\%, as both benchmarks require more audio information to answer questions. When $r$ increases, performance on all benchmarks went up until $r=5$ where further increasing audio token suppresses visual memory, and the gain obtained from larger audio memory is insufficient to balance out the loss form visual memory reduction.

\subsection{Runtime Memory and Latency}
\begin{figure}[h]
    \centering
    \includegraphics[width=\linewidth]{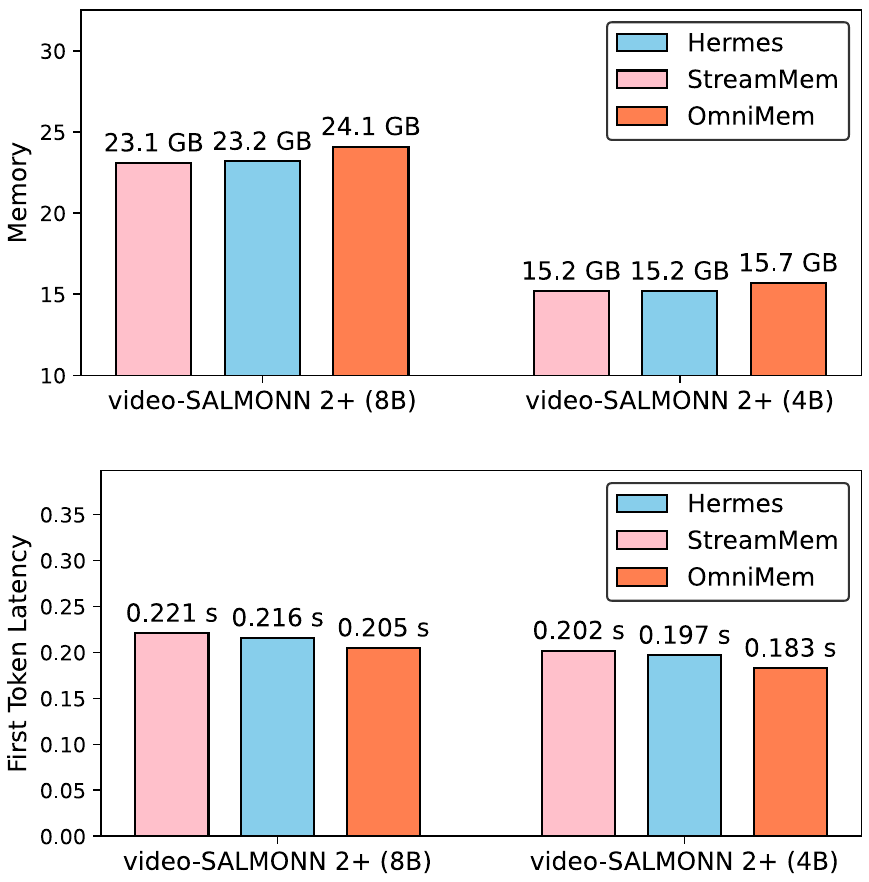}
    \caption{Runtime statistics, including peak memory consumption and first token latency using video-SALMONN 2+ (8B) and (4B) on a single H800 GPU. 8K per-layer memory size is used.}
    \label{fig:runtime}
\end{figure}

We plot the runtime statistics, including the memory and first-token latency (the time taken from receiving the user query to generating the first token), as shown in Fig. \ref{fig:runtime}.

All 3 methods consumed about the same level of GPU memory, with a slight $<1$GB overhead in OmniMem. This overhead was mainly due to the retain of hidden states for cosine similarity redundancy computation, which we found important to achieve good performance for videos over 1 hour. This overhead purely depends on the memory size and model size, hence more negligible for the 4B model. The dominant memory cost for all methods is from the model parameters.

Regarding first-token latency, all three methods achieved similar latency, which is around 0.2 seconds and does not change much across different models sizes. Moreover, we found that the non-uniform per-layer budget allocation slightly improved the latency over uniform baselines. 

\subsection{Sensitivity to Hyper-parameters}

\begin{figure*}[t]
    \centering
    \includegraphics[width=\linewidth]{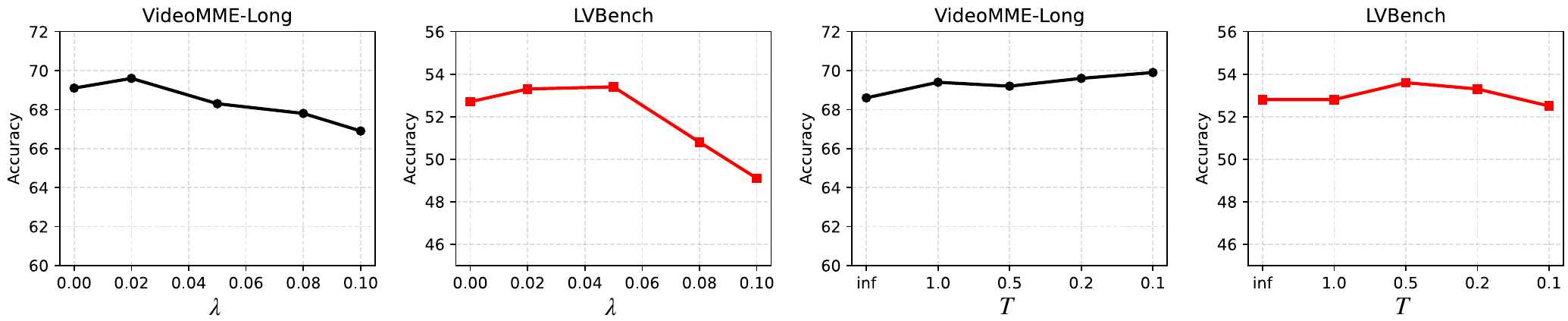}
    \caption{Sensitivity to the key hyper-parameters $\lambda$ in Eqn. \eqref{eq:psi}, and temperature $T$ in Eqn. \eqref{eq:alloc}. Experiments are performed using the video-SALMONN 2+ (8B) model on VideoMME long partition and LVBench datasets.}
    \label{fig:sensitivity}
\end{figure*}

\begin{figure*}[t]
    \centering
    \includegraphics[width=\linewidth]{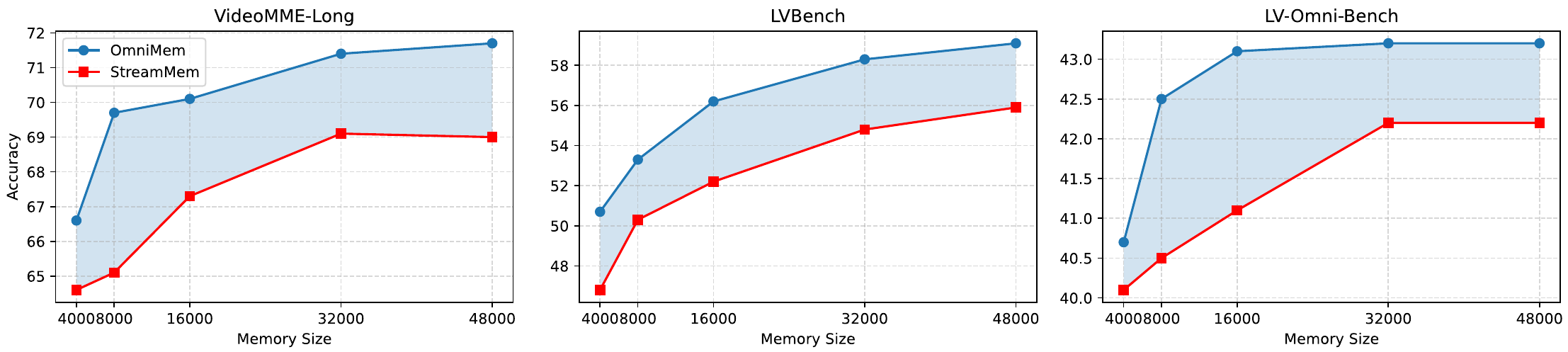}
    \caption{Accuracy variation against varying memory size from 4k to 48k per layer using video-SALMONN 2+ (8B) on VideoMME-long, LVBench and LV-Omni-Bench for both OmniMem (ours) and StreamMem.}
    \label{fig:memvar}
\end{figure*}

There are two key hyper-parameters included in OmniMem that needs tuning, including the $\lambda$ in Eqn. \eqref{eq:psi} which determines the contribution of the attention score to the overall selection score, and the temperature $T$ in Eqn. \eqref{eq:alloc} determining the variation of budget sizes across Transformer layers. We vary the two hyper-parameters within reasonable ranges and measure the change in model performance on VideoMME long partition and LVBench, as shown in Fig. \ref{fig:sensitivity}.

\textbf{Accuracy is relatively stable for varying $\lambda$} for $\lambda < 0.1$. While we picked 0.02 for $\lambda$, the performance all the way up to $\lambda=0.05$ is relatively stable, and mostly better than $\lambda=0.0$. This demonstrate the usefulness of the attention weights as a signal to flag which tokens are used more often by the next layer. However, at higher $\lambda$ values when attention score dominant, we found a large performance drop, indicating importance score alone is far from being enough, and redundancy measures (i.e. the cosine similarity) is necessary.

\textbf{Accuracy is stable for $T$} for a reasonable range, e.g. $T>0.1$. In general, using a variable layerwise budget achieves better performance. The layerwise budget assigns larger sizes to earlier layers and smaller sizes to later layers. Within this range, the accuracy varies within 2\% for both VideoMME long and LVBench. We believe this stability partly comes from the maximum-minimum value restrictions we applied on each layer.

\subsection{Influence of Memory Size}

We show the variation of model accuracy against different memory budget sizes from 4k to 48k per layer using video-SALMONN 2+ (8B) on the 3 long video understanding benchmarks as shown in Fig. \ref{fig:memvar}. We can go up to 48k because the maximum sequence length that video-SALMONN 2+ has been trained on exceeds 50k.

\textbf{OmniMem consistently performs better under different budgets} compared to StreamMem with a consistent and clear margin as shown in Fig. \ref{fig:memvar}. The advantage is at its largest for medium-sized budgets (e.g. 8k or 16k). The performance plateaus when memory size exceeds 32k for Video-MME long and LV-Omni-Bench, but still shows an upward trend for LVBench. 

\textbf{Visual to audio ratio becomes more important at a smaller budget, e.g. 4k}. Under that budget, a ratio $r=5$ leaves around only 3k vectors for visual, and the visual budget is far from being enough for long videos. At this size, whether using 800 ($r=5$) or 400 ($r=10$) vectors for audio cannot compensate for the large information loss in visual modality, and hence the difference between OmniMem and StreamMem becomes smaller, although the former still outperforms the latter.

\section{Conclusion}

This paper presented OmniMem, a perturbation-aware, modality-aware memory compression method for chunked audio-visual streaming. By combining attention importance, redundancy reduction, and adaptive audio-visual budget allocation, OmniMem preserves the most informative KV pairs under limited memory. Experiments on VideoMME Long, LVBench, and LVOmniBench show consistent 2–3\% accuracy gains over strong baselines, with further improvements from budget-aware fine-tuning, demonstrating its effectiveness for efficient long-form audio-visual understanding.

\section{Limitations}

One of the limitations is the memory overhead caused by the retention of hidden states for cosine similarity computation. We believe better implementation or memory management could be done to fix that or keep that to a negligible level. We also have a limitation in our experiments which was caused by the shortage of streaming benchmarks that require both audio and visual inputs, especially in the long video understanding direction. Last, this method may need modification or re-design of the selection criterion to function better on linear-attention-based Transformer models.


\begin{figure*}[t]
    \centering
    \includegraphics[width=\linewidth]{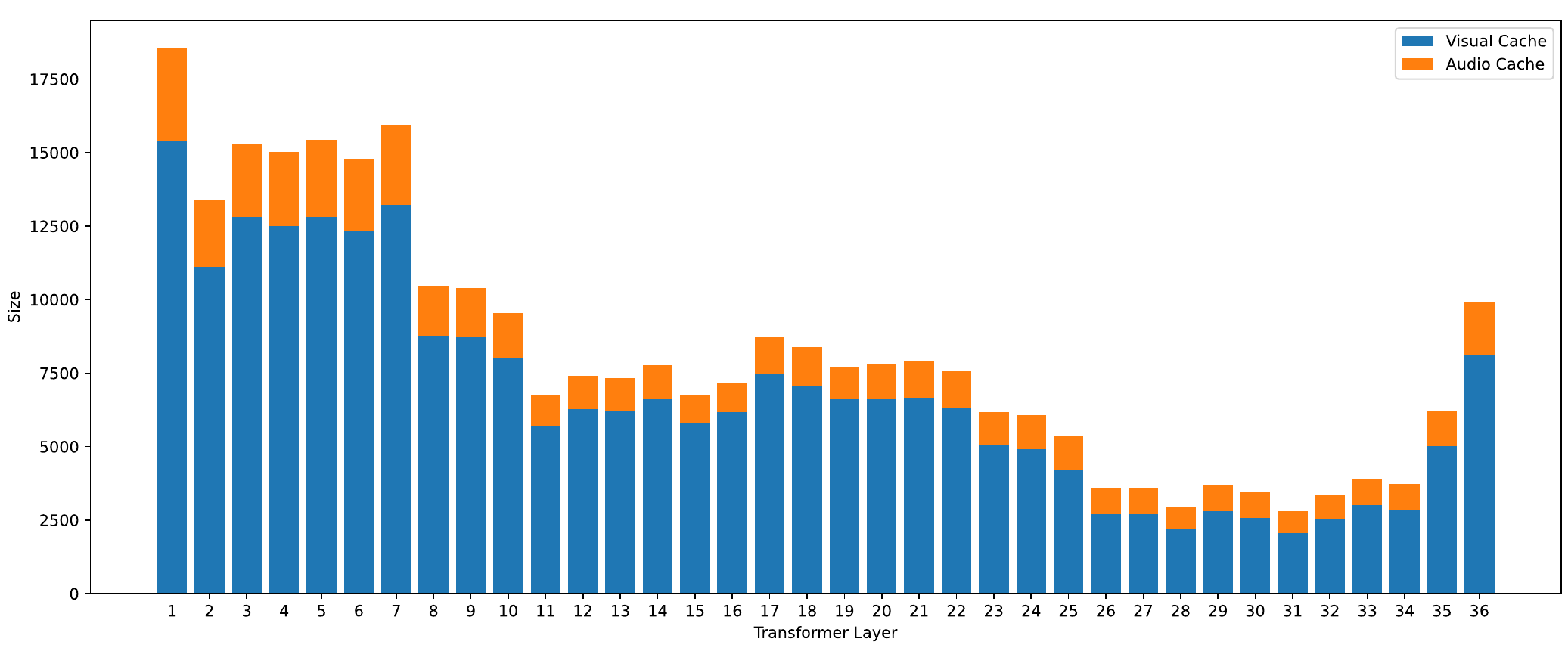}
    \caption{Budget distribution across layers for video-SALMONN 2+ (8B) with $T=0.2$.}
    \label{fig:budget_distribution}
\end{figure*}

\begin{table*}[t]
    \centering
        \caption{Ablation studies on core components of OmniMem using video-SALMONN 2+ (8B) model. When not using separate audio visual budgets, we still keep the same overall budget allocation per layer unchanged (which is non-uniform). When not using $\psi_k$, we use cosine similarity alone as the selection criterion.}
    \resizebox{0.8\linewidth}{!}{
    \begin{tabular}{lcccc}
    \toprule
    Methods     & Video-MME long & LVBench & LV-Omni-Bench  \\
    \midrule
    OmniMem     & 69.6 & 53.3 & 42.5 \\
    ~~~~ w/o Separate audio visual budgets & 66.8 & 52.2 & 40.7 \\
    ~~~~ w/o AVBA (i.e. uniform single cache) & 66.3 & 51.7 & 40.5  \\
    ~~~~ w/o $\psi_k$ & 67.9 & 52.0 & 41.4 \\
    \bottomrule
    \end{tabular}}
    \label{tab:ablation}
\end{table*}

\bibliography{custom}

@article{streammem,
  title         = {StreamMem: Query-Agnostic KV Cache Memory for Streaming Video Understanding},
  author        = {Yanlai Yang and Zhuokai Zhao and Satya Narayan Shukla and Aashu Singh and Shlok Kumar Mishra and Lizhu Zhang and Mengye Ren},
  journal       = {arXiv:2508.15717},
  year          = {2025},
}

@article{streamingbench,
      title={StreamingBench: Assessing the Gap for MLLMs to Achieve Streaming Video Understanding}, 
      author={Junming Lin and Zheng Fang and Chi Chen and Zihao Wan and Fuwen Luo and Peng Li and Yang Liu and Maosong Sun},
      year={2024},
      journal={arXiv:2411.03628},
}

@article{streamkv,
  title         = {StreamKV: Streaming Video Question-Answering with Segment-based KV Cache Retrieval and Compression},
  author        = {Yilong Chen and Xiang Bai and Zhibin Wang and Chengyu Bai and Yuhan Dai and Ming Lu and Shanghang Zhang},
  journal       = {arXiv:2511.07278},
  year          = {2025},
}

@inproceedings{infinipotv,
  title         = {InfiniPot-V: Memory-Constrained KV Cache Compression for Streaming Video Understanding},
  author        = {Minsoo Kim and Kyuhong Shim and Jungwook Choi and Simyung Chang},
  booktitle       = {Proc. NeurIPS 2025},
  year          = {2025},
}

@inproceedings{hermes,
    title={HERMES: KV Cache as Hierarchical Memory for Efficient Streaming Video Understanding}, 
      author={Haowei Zhang and Shudong Yang and Jinlan Fu and See-Kiong Ng and Xipeng Qiu},
    year={2026},
  booktitle       = {Proc. ACL 2026},
}

@article{videosalmonn2,
  title         = {video-SALMONN 2: Caption-Enhanced Audio-Visual Large Language Models},
  author        = {Tang, Changli and Li, Yixuan and Yang, Yudong and Zhuang, Jimin and Sun, Guangzhi and Li, Wei and Ma, Zejun and Zhang, Chao},
  journal       = {arXiv:2506.15220},
  year          = {2025},
}

@article{qwen25omni,
  title         = {Qwen2.5-Omni Technical Report},
  author        = {Jin Xu and Zhifang Guo and Jinzheng He and Hangrui Hu and Ting He and Shuai Bai and Keqin Chen and Jialin Wang and Yang Fan and Kai Dang and Bin Zhang and Xiong Wang and Yunfei Chu and Junyang Lin},
  journal       = {arXiv:2503.20215},
  year          = {2025},
}

@article{videomme,
  title         = {Video-MME: The First-Ever Comprehensive Evaluation Benchmark of Multi-modal LLMs in Video Analysis},
  author        = {Chaoyou Fu and Yuhan Dai and Yongdong Luo and Lei Li and Shuhuai Ren and Renrui Zhang and Zihan Wang and Chenyu Zhou and Yunhang Shen and Mengdan Zhang and Peixian Chen and Yanwei Li and Shaohui Lin and Sirui Zhao and Ke Li and Tong Xu and Xiawu Zheng and Enhong Chen and Caifeng Shan and Ran He and Xing Sun},
  journal       = {arXiv:2405.21075},
  year          = {2024},
}

@article{lvbench,
  title         = {LVBench: An Extreme Long Video Understanding Benchmark},
  author        = {Weihan Wang and Zehai He and Wenyi Hong and Yean Cheng and Xiaohan Zhang and Ji Qi and Xiaotao Gu and Shiyu Huang and Bin Xu and Yuxiao Dong and Ming Ding and Jie Tang},
  journal       = {arXiv:2406.08035},
  year          = {2024},
}

@article{lvomnibench,
  title         = {LVOmniBench: Pioneering Long Audio-Video Understanding Evaluation for Omnimodal LLMs},
  author        = {Keda Tao and Yuhua Zheng and Jia Xu and Wenjie Du and Kele Shao and Hesong Wang and Xueyi Chen and Xin Jin and Junhan Zhu and Bohan Yu and Weiqiang Wang and Jian Liu and Can Qin and Yulun Zhang and Ming-Hsuan Yang and Huan Wang},
  journal       = {arXiv:2603.19217},
}

@inproceedings{pemf,
      title={StreamForest: Efficient Online Video Understanding with Persistent Event Memory}, 
      author={Xiangyu Zeng and Kefan Qiu and Qingyu Zhang and Xinhao Li and Jing Wang and Jiaxin Li and Ziang Yan and Kun Tian and Meng Tian and Xinhai Zhao and Yi Wang and Limin Wang},
      year={2025},
      booktitle={Proc. NeurIPS}
}

@inproceedings{vss,
      title={video-SALMONN S: Memory-Enhanced Streaming Audio-Visual LLM}, 
      author={Guangzhi Sun and Yixuan Li and Xiaodong Wu and Yudong Yang and Wei Li and Zejun Ma and Chao Zhang},
      year={2026},
      booktitle={Proc. ICML}
}

@article{pyramid,
      title={PyramidKV: Dynamic KV Cache Compression based on Pyramidal Information Funneling}, 
      author={Zefan Cai and Yichi Zhang and Bofei Gao and Yuliang Liu and Yucheng Li and Tianyu Liu and Keming Lu and Wayne Xiong and Yue Dong and Junjie Hu and Wen Xiao},
      year={2025},
      journal={arXiv: 2406.02069},
}

@article{squeezed,
  title={Squeezed Attention: Accelerating Long Context Length LLM Inference},
  author={Hooper, Coleman and Kim, Sehoon and Mohammadzadeh, Hiva and Maheswaran, Monishwaran and Paik, June and Mahoney, Michael W and Keutzer, Kurt and Gholami, Amir},
  journal={arXiv:2411.09688},
  year={2024}
}

@article{lava,
      title={LAVa: Layer-wise KV Cache Eviction with Dynamic Budget Allocation}, 
      author={Yiqun Shen and Song Yuan and Zhengze Zhang and Xiaoliang Wang and Daxin Jiang and Nguyen Cam-Tu},
      year={2025},
      journal={arXiv:2509.09754},
}

@article{evolkv,
      title={EvolKV: Evolutionary KV Cache Compression for LLM Inference}, 
      author={Bohan Yu and Yekun Chai},
      year={2025},
      journal={arXiv:2509.08315},
}

@article{rekv,
  author       = {Shangzhe Di and Zhelun Yu and Guanghao Zhang and Haoyuan Li and Tao Zhong and Hao Cheng and Bolin Li and Wanggui He and Fangxun Shu and Hao Jiang},
  title        = {Streaming Video Question-Answering with In-Context Video KV-Cache Retrieval},
  journal      = {arXiv preprint arXiv:2503.00540},
  year         = {2025}
}

@article{longva,
  title={Long Context Transfer from Language to Vision},
  author={Peiyuan Zhang and Kaichen Zhang and Bo Li and Guangtao Zeng and Jingkang Yang and Yuanhan Zhang and Ziyue Wang and Haoran Tan and Chunyuan Li and Ziwei Liu},
  journal={arXiv preprint arXiv:2406.16852},
  year={2024},
}

@article{longvu,
    title={{LongVU}: {S}patiotemporal Adaptive Compression for Long Video-Language Understanding},
    author={Shen, Xiaoqian and Xiong, Yunyang and Zhao, Changsheng and Wu, Lemeng and Chen, Jun and Zhu, Chenchen and Liu, Zechun and Xiao, Fanyi and Varadarajan, Balakrishnan and Bordes, Florian and Liu, Zhuang and Xu, Hu and J. Kim, Hyunwoo and Soran, Bilge and Krishnamoorthi, Raghuraman and Elhoseiny, Mohamed and Chandra, Vikas},
    journal={arXiv:2410.17434},
    year={2024}
  }

@article{shu2024video,
  title={{Video-XL}: {E}xtra-Long Vision Language Model for Hour-Scale Video Understanding},
  author={Shu, Yan and Zhang, Peitian and Liu, Zheng and Qin, Minghao and Zhou, Junjie and Huang, Tiejun and Zhao, Bo},
  journal={arXiv preprint arXiv:2409.14485},
  year={2024}
}

@inproceedings{streamingtom,
  title={StreamingTOM: Streaming Token Compression for Efficient Video Understanding},
  author={Chen, Xueyi and Tao, Keda and Shao, Kele and Wang, Huan},
  booktitle={CVPR},
  year={2026}
}

@inproceedings{moviechat,
      title={{MovieChat}: From Dense Token to Sparse Memory for Long Video Understanding}, 
      author={Enxin Song and Wenhao Chai and Guanhong Wang and Yucheng Zhang and Haoyang Zhou and Feiyang Wu and Haozhe Chi and Xun Guo and Tian Ye and Yanting Zhang and Yan Lu and Jenq-Neng Hwang and Gaoang Wang},
      year={2024},
      booktitle={Proc. CVPR}
}

@article{longvila,
      title={LongVILA: Scaling Long-Context Visual Language Models for Long Videos},
      author={Yukang Chen and Fuzhao Xue and Dacheng Li and Qinghao Hu and Ligeng Zhu and Xiuyu Li and Yunhao Fang and Haotian Tang and Shang Yang and Zhijian Liu and Yihui He and Hongxu Yin and Pavlo Molchanov and Jan Kautz and Linxi Fan and Yuke Zhu and Yao Lu and Song Han},
      year={2024},
      eprint={2408.10188},
      archivePrefix={arXiv},
      primaryClass={cs.CV}
}

@article{liu2025video,
  title={{Video-XL-Pro}: {R}econstructive Token Compression for Extremely Long Video Understanding},
  author={Liu, Xiangrui and Shu, Yan and Liu, Zheng and Li, Ao and Tian, Yang and Zhao, Bo},
  journal={arXiv preprint arXiv:2503.18478},
  year={2025}
}

\appendix

\section{Layer-level Budget Distribution}
\label{sec:distribution}
The distribution of budget allocation across different layers is shown in Fig. \ref{fig:budget_distribution}

\section{Guidance Prompts}
We use the following guidance prompt for StreamMem and HERMES.

\small \begin{promptbox}[label={real_scripts}]{Guidance Prompt}

Summarize the video narrative, identifying main characters, key events, timeline changes, and the overall
theme. Pay attention to both audio and visual content.

\end{promptbox}

\section{Ablation Studies}
\label{sec:ablation}

We provide ablation studies for the two key components: AVBA and scoring $\psi_k$, in Table \ref{tab:ablation}. When not using $\psi_k$, we just simply use cosine similarity as the selection criterion. Removing AVBA or removing the separate allocation to audio cache has a big impact to the performance, especially for video-MME long partition where audio plays a more important role in understanding. 

For the case with only a single cache, a slight degradation in performance was found on long video understanding benchmarks when a uniform per-layer budget was used, compared to a layer-dependent budget derived from the entropy and similarity from Eqn. \eqref{eq:alloc}. Removing $\psi_k$ and only using cosine similarity for selection also degrades the performance, but its influence is rather modest compared to the AVBA module.

\end{document}